\documentclass[11pt,a4paper]{article}
\usepackage{acl2015}
\usepackage{times}
\usepackage{url}
\usepackage{latexsym}
\usepackage{amsmath,bm}
\usepackage{amsfonts}
\usepackage{mathrsfs}
\usepackage{amsbsy}
\usepackage{graphicx}
\usepackage{xcolor}
\usepackage{multirow}
\usepackage{paralist}
\usepackage{subfigure}

\title{A Comparative Study on Regularization Strategies\\ for Embedding-based Neural Networks}
\author{Hao Peng\thanks{\ \ Equal contribution.\quad $^\dag$Corresponding author.},\,$^1$\;
Lili Mou,\!$^{*1}$
Ge Li,\!$^{\dag1}$\ \ Yunchuan Chen,$^2$ Yangyang Lu,$^1$ Zhi Jin$^1$\\
$^1$Software Institute, Peking University, 100871, P. R. China\\
\{penghao.pku, doublepower.mou\}@gmail.com,\{lige, luyy11, zhijin\}@sei.pku.edu.cn\\
$^2$University of Chinese Academy of Sciences, chenyunchuan11@mails.ucas.ac.cn
}
\date{}
\begin{document}
\maketitle
\begin{abstract}
This paper aims to compare different regularization
strategies to address a common phenomenon, severe overfitting, in embedding-based neural networks for NLP.
We chose two widely studied neural models and tasks as our testbed.
We tried several frequently applied or newly proposed regularization strategies, including
penalizing weights (embeddings excluded), penalizing embeddings, re-embedding words, and
dropout. We also emphasized on incremental hyperparameter tuning, and combining
different regularizations. The results provide a picture on tuning
hyperparameters for neural NLP models.

\end{abstract}

\section{Introduction}

Neural networks have exhibited considerable potential in various fields \cite{imagenet,speechrnn}. 
In early years on neural NLP research, neural networks were used in language modeling \cite{LM,hierarchical,hierarchical2};
recently, they have been applied to various supervised tasks, such as named entity recognition \cite{unified}, sentiment analysis \cite{RAE,sentenceTBCNN}, relation classification \cite{re,sdplstm}, etc.
In the field of NLP, neural networks are typically combined with word embeddings, which are usually first pretrained
by unsupervised algorithms like \newcite{word2vec}; then they are fed forward to standard neural models, fine-tuned during supervised learning. However, embedding-based neural networks usually suffer from severe overfitting because of the high dimensionality 
of parameters.

A curious question is whether we can regularize embedding-based NLP neural models to improve
generalization.
Although existing and newly proposed regularization methods might alleviate
the problem, their inherent performance in neural NLP models is not clear: 
the use of embeddings is sparse; the behaviors may be different from those in other scenarios like image recognition.
Further, selecting hyperparameters to pursue the best performance
by validation is extremely time-consuming, as suggested in \newcite{scratch}. Therefore, new studies are needed to provide a more
complete picture regarding regularization for neural natural language processing. 
Specifically, we focus on the following research questions in this paper.
\begin{compactitem}
\item[RQ 1:] How do different regularization strategies typically behave in embedding-based neural networks?
\item[RQ 2:] Can regularization coefficients be tuned incrementally during training so as to ease the burden of hyperparameter tuning?
\item[RQ 3:] What is the effect of combining different regularization strategies?
\end{compactitem}

In this paper, we systematically and quantitatively
compared four different regularization strategies,
namely penalizing weights, penalizing embeddings, newly proposed word re-embedding \cite{reembed},
and dropout \cite{dropout}.
We analyzed these regularization methods by two widely studied models and tasks.
We also emphasized on incremental hyperparameter tuning and the combination of different regularization methods.

Our experiments provide some interesting results:
(1) Regularizations do help generalization, but their effect depends largely
on the datasets' size.
(2) Penalizing $\ell_2$-norm of embeddings helps optimization as well,
improving training accuracy unexpectedly.
(3) Incremental hyperparameter tuning achieves similar performance, indicating
that regularizations mainly serve as a ``local'' effect.
(4) Dropout performs slightly worse than $\ell_2$ penalty in our experiments; 
however, provided very small $\ell_2$ penalty, dropping out hidden units and 
penalizing $\ell_2$-norm are generally complementary.
(5) The newly proposed re-embedding words method is not effective in our experiments.

\vspace{-.1cm}
\section{Tasks, Models, and Setup}\label{sModelTask}
\vspace{-.2cm}
\textbf{Experiment I: Relation extraction}. The dataset in this experiment comes from SemEval-2010 Task 8.\footnote{
http://www.aclweb.org/anthology/S10-1006
}
The goal is to classify the relationship between two marked entities in each sentence.
We refer interested readers to recent advances, e.g., \newcite{re0},
\newcite{re}, and \newcite{sdplstm}.
To make our task and model general, however,
we do not consider entity tagging information; we do not distinguish the order of two entities either. In total, there are 10 labels, i.e., 9 different relations plus a default \verb|other|.

Regarding the neural model, we applied Collobert's
convolutional neural network (CNN) \cite{unified} with minor modifications. 
The model comprises a fixed-window convolutional layer with size equal to 5, $\bm 0$ padded at the end of each sentence; 
a max pooling layer; a $\tanh$ hidden layer;
and a $\operatorname{softmax}$ output layer.

\textbf{Experiment II: Sentiment analysis}. This is another testbed for neural NLP, aiming to predict the sentiment
of a sentence. The dataset is the Stanford sentiment treebank \cite{RAE}\footnote{
http://nlp.stanford.edu/sentiment/
}; target labels are \verb|strongly/weakly| \verb|positive/negative|, or \verb|neutral|.

We used the recursive neural network (RNN), which is proposed in \newcite{RAE}, and further developed in \newcite{matrixvector}; \newcite{deepRNN}. 
RNNs make use of binarized constituency trees, and
recursively encode children's information to their parent's; 
the root vector is finally used for sentiment classification.

\textbf{Experimental Setup}. To setup a fair comparison, 
we set all layers to be 50-dimensional in advance (rather than by validation). Such setting has been used in previous work like \newcite{adahsm}.
Our embeddings are pretrained on the Wikipedia corpus using \newcite{unified}.
The learning rate is 0.1 and fixed in Experiment I; for RNN, however,
we found learning rate decay 
helps to prevent parameter blowup (probably due to the recursive, and thus chaotic nature).
Therefore, we applied power decay \cite{learningrate} with power equal to $-1$.
For each strategy, we tried a large range of regularization coefficients, $10^{-9}, \cdots, 10^{-2}$, extensively
from underfitting to no effect with granularity 10x. We ran the model 5 times with different initializations.
We used mini-batch stochastic gradient descent;
gradients are computed by standard backpropagation. 
For source code, please refer to our project website.\footnote{
https://sites.google.com/site/regembeddingnn/
}

It needs to be noticed that,
the goal of this paper is not to outperform or reproduce state-of-the-art results.
Instead, we would like to have a fair comparison. 
The testbed of our work is two widely studied models and tasks,
which were not chosen on purpose. 
During the experiments, we tried to make the comparison as fair as possible.
Therefore, we think that the results of this work can be generalized
to similar scenarios.

\begin{figure*}[!t]
\begin{center}
${\includegraphics[width=.43\textwidth]{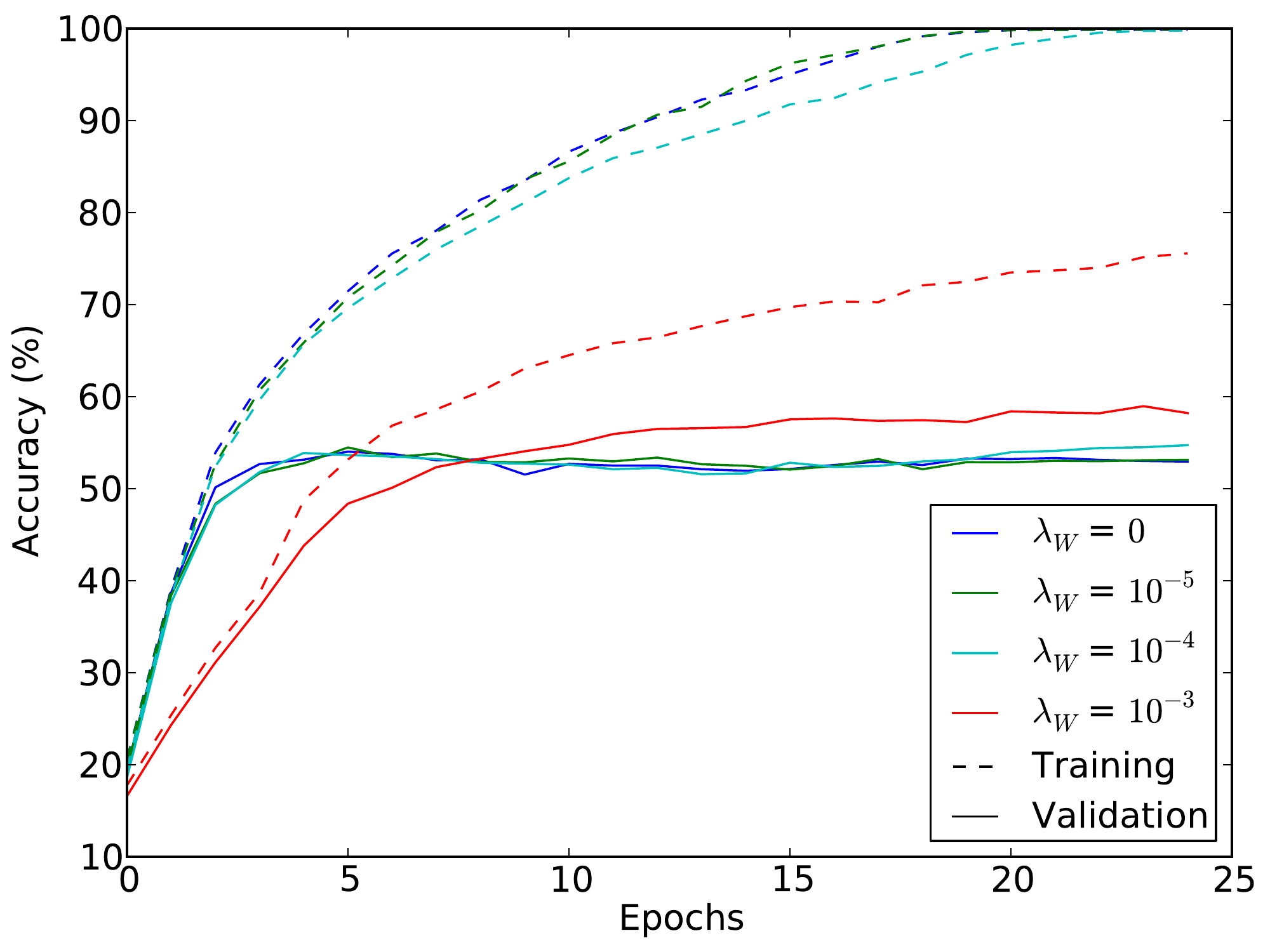}\atop \text{(a) Penalizing
weights in Experiment I.}}$\ \ \ \ \ \ 
${\includegraphics[width=.43\textwidth]{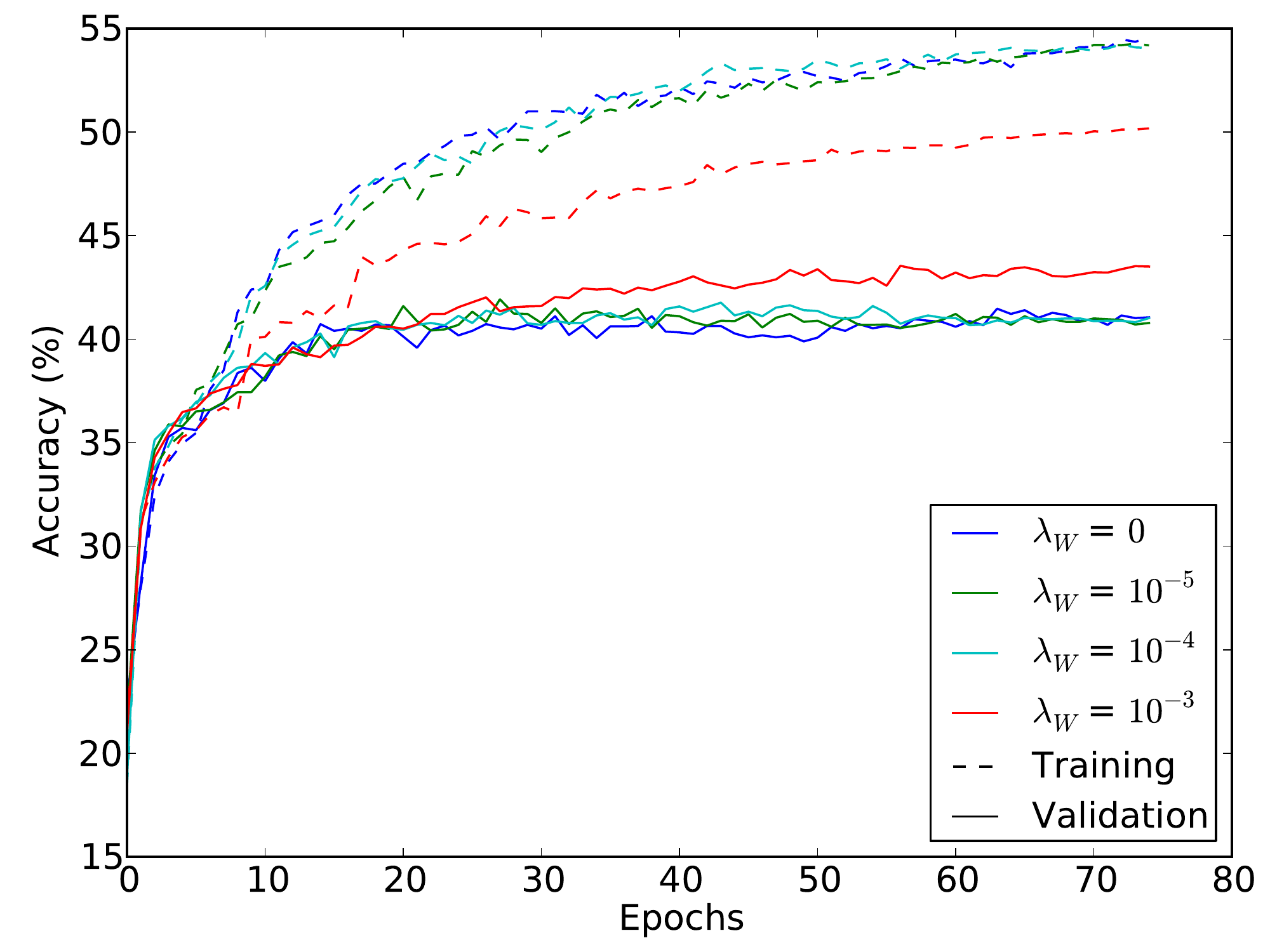}\atop \text{(b) Penalizing
weights in Experiment II.}}$\\

${\includegraphics[width=.43\textwidth]{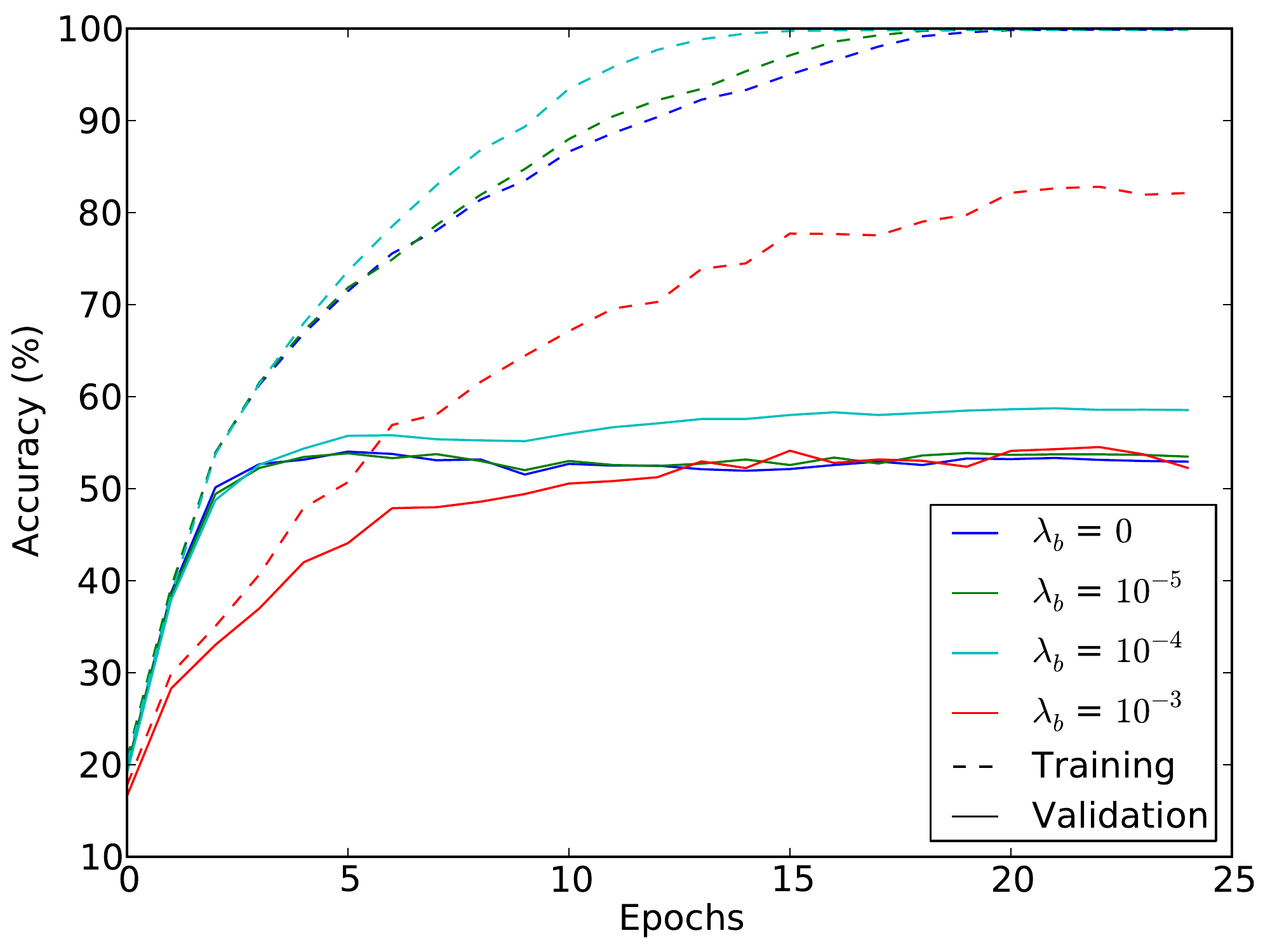}\atop\text{(c) Penalizing
embeddings in Experiment I.}}$\ \ \ \ \ \ 
${\includegraphics[width=.43\textwidth]{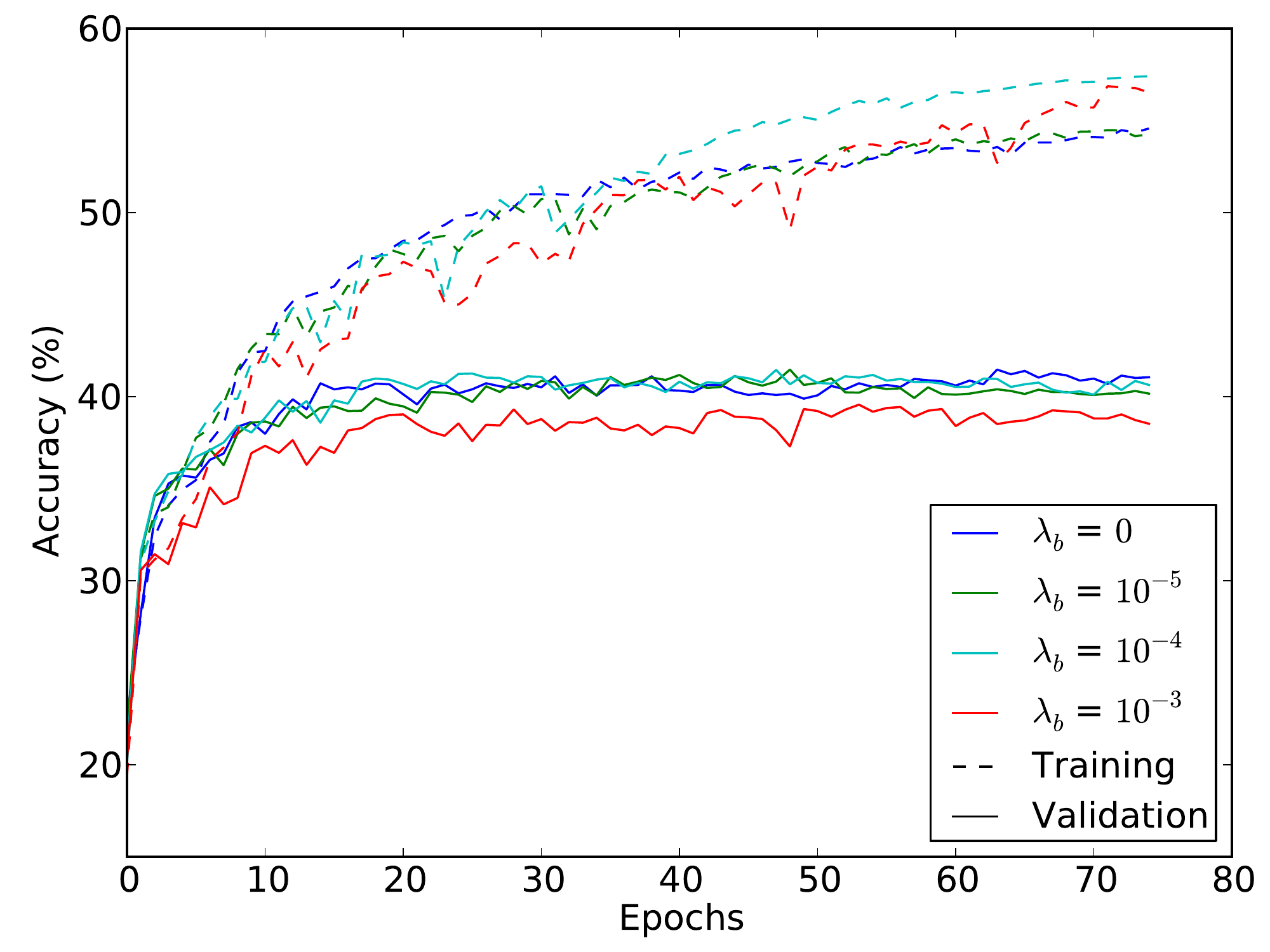}\atop\text{(d) Penalizing
embeddings in Experiment II.}}$

${\includegraphics[width=.43\textwidth]{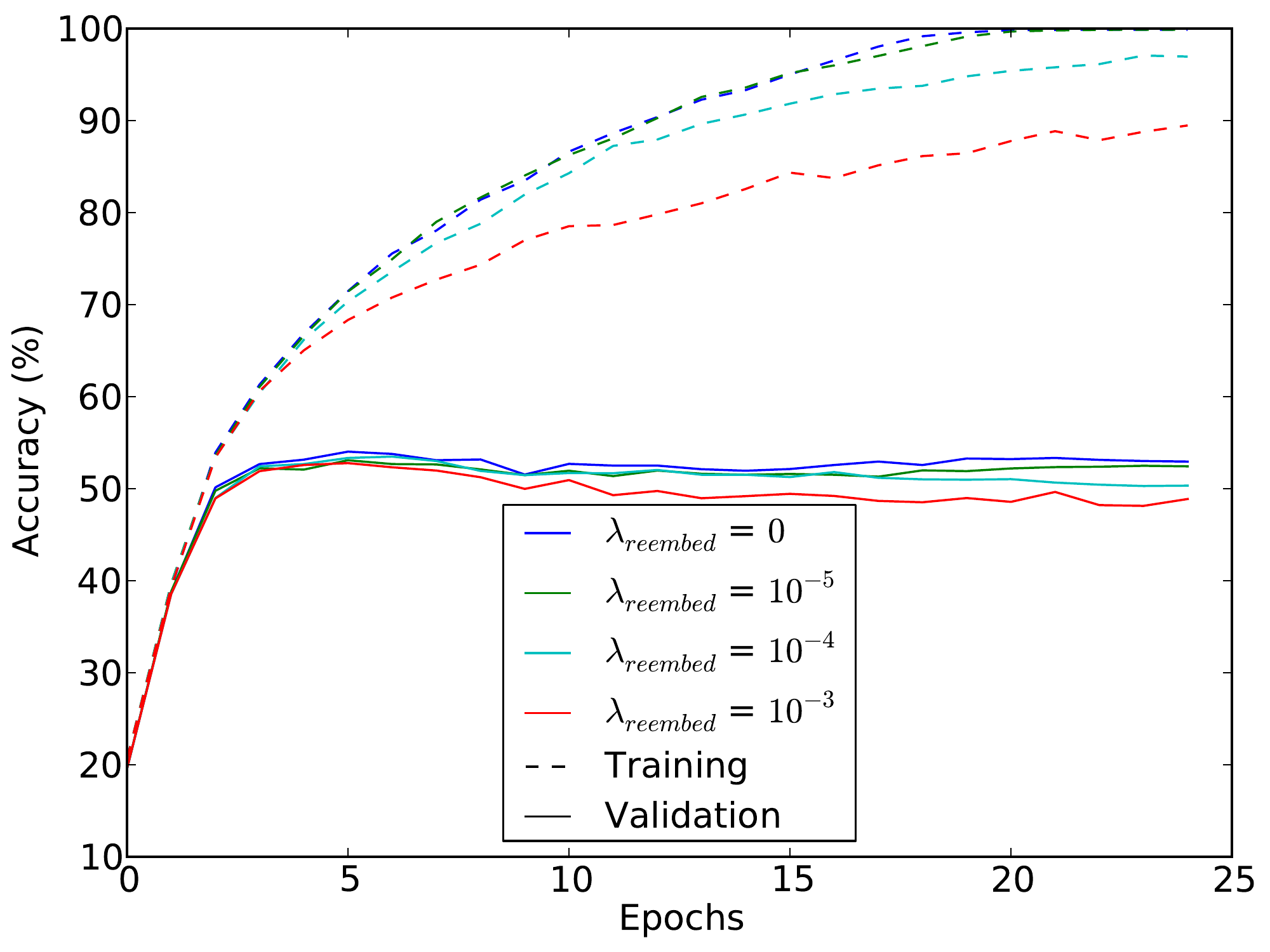}\atop\text{(e) Re-embedding words in Experiment I.}}$\ \ \ \ \ \ 
${\includegraphics[width=.43\textwidth]{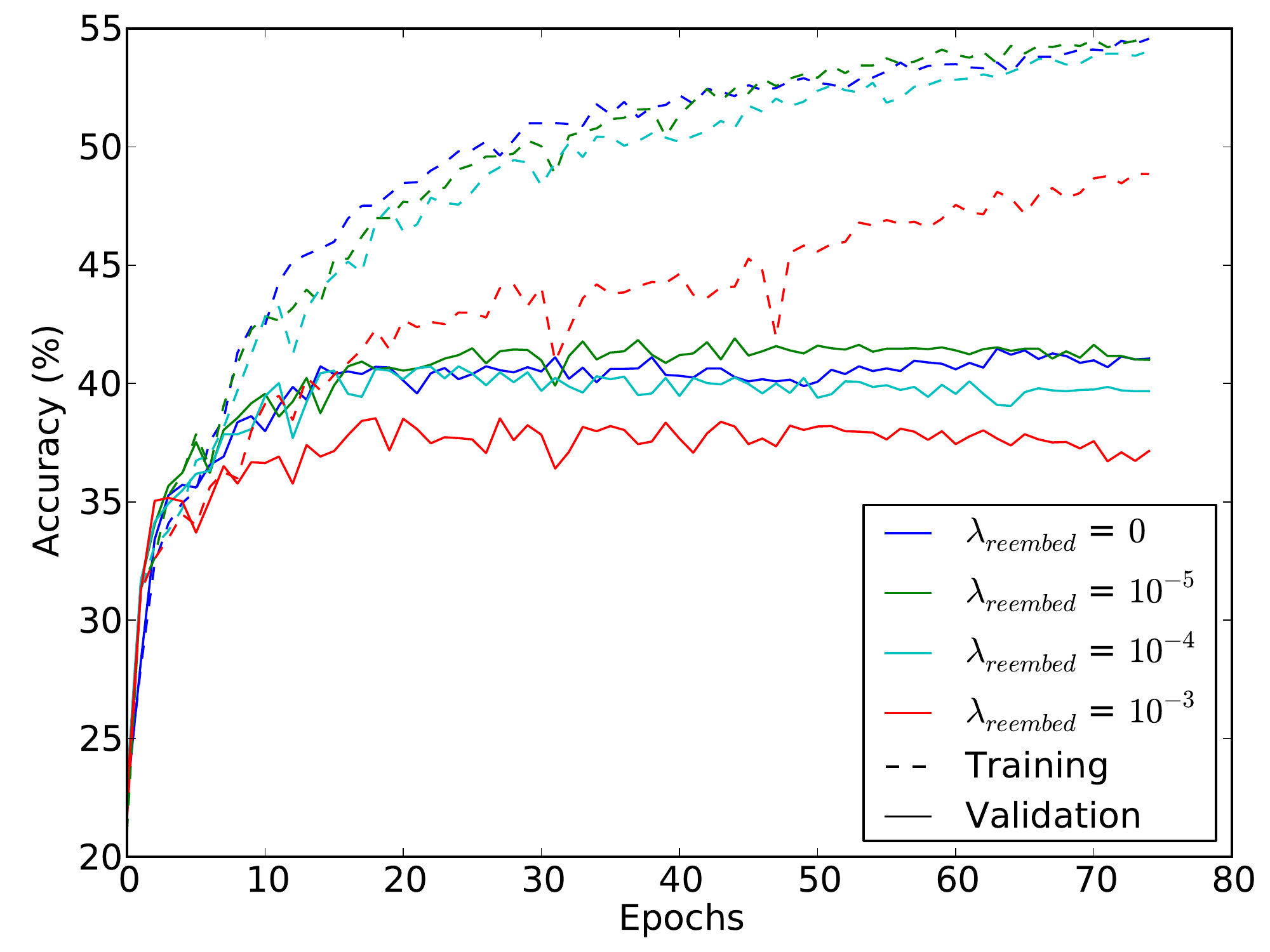}\atop\text{(f) Re-embedding words in Experiment II.}}$

$\ \ {\includegraphics[width=.43\textwidth]{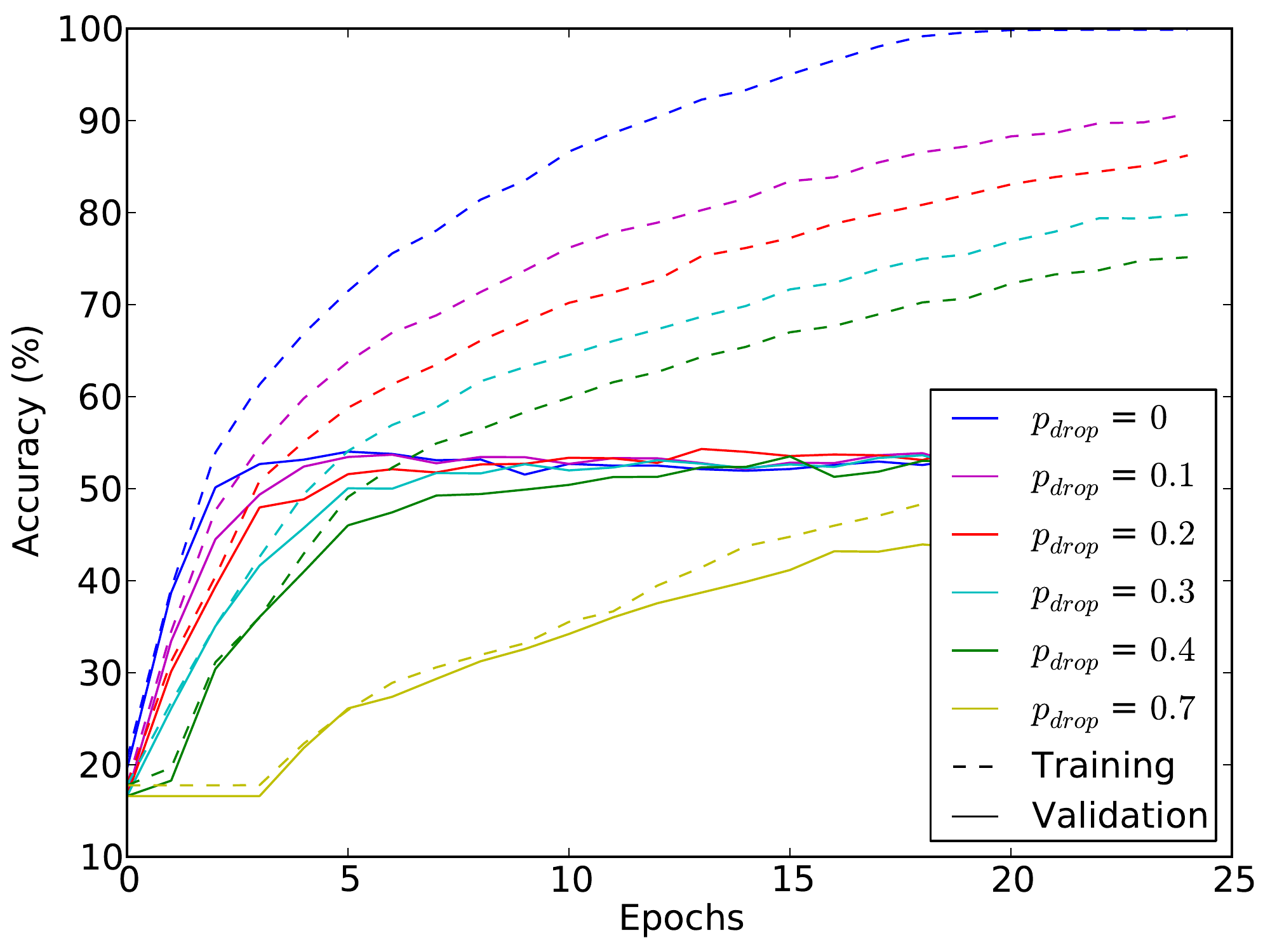}
\atop
{
     \text{\footnotesize (g) Applying dropout in Experiment I. $p=0.5,0.6$ }
\atop
    \text{\footnotesize are omitted because they are similar to small values.}
} 
}$\ \ \ \ \ \ 
${\includegraphics[width=.43\textwidth]{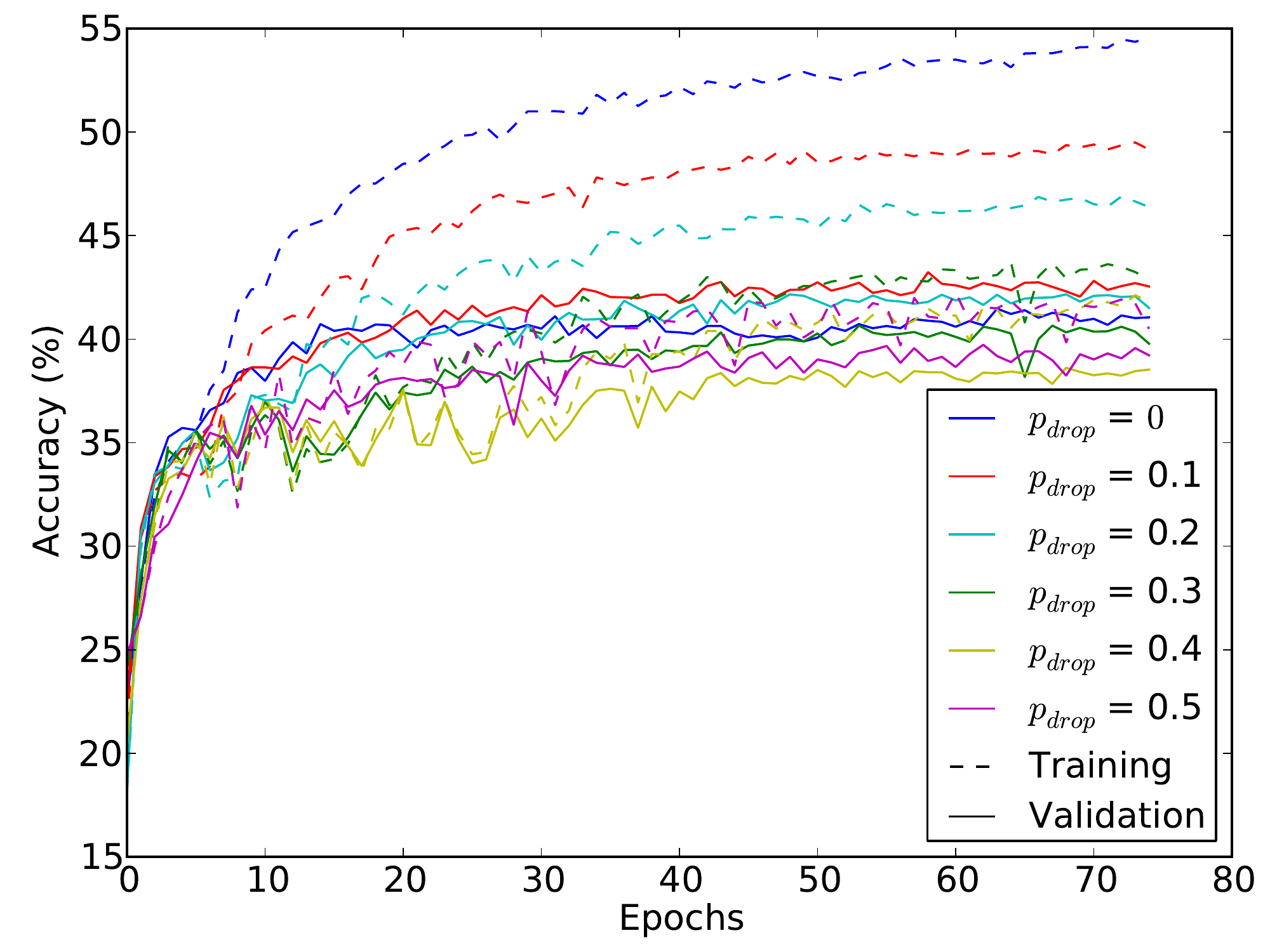}\atop\text{(h) Applying dropout in Experiment II.}}$

\caption{\small Averaged learning curves. Left: Experiment I, relation extraction with CNN. Right: Experiment II, sentiment analysis with RNN. From top to bottom, we penalize weights, penalize embeddings, re-embed words, and drop out. Dashed lines refer to training accuracies; solid lines are validation accuracies.}
\label{fIndividual}
\end{center}
\end{figure*}

\vspace{-.1cm}
\section{Regularization Strategies}\label{sReg}
\vspace{-.2cm}
In this section, we describe four regularization strategies used in our experiment.

\begin{compactitem}
\item Penalizing $\ell_2$-norm of weights. Let $E$ be the cross-entropy error for classification, and
$R$ be a regularization term. The overall cost function is 
$J=E+\lambda R$, 
where $\lambda$ is the coefficient.  In this case, $R=\|W\|^2$, and the coefficient is denoted as $\lambda_W$.
\item Penalizing $\ell_2$-norm of embeddings. 
Some studies do not distinguish embeddings or connectional weights for regularization
\cite{lstm1}. However, we would like to analyze their effect separately, for embeddings are sparse in use. Let $\Phi$ denote embeddings; then we have $R = \|\Phi\|^2$. 
\item Re-embedding words \cite{reembed}.
Suppose $\Phi_0$ denotes the original embeddings trained on a large corpus, and $\Phi$ denotes the embeddings fine-tuned during supervised training.
We would like to penalize the norm of the difference between $\Phi_0$ and $\Phi$, i.e.,
$R = \|\Phi_0-\Phi\|^2$. 
In the limit of penalty to infinity, the model is mathematically equivalent to
``frozen embeddings,'' where word vectors are used as surface features.

\item Dropout \cite{dropout}. In this strategy, each neural node
is set to 0 with a predefined dropout probability $p$ during training; when testing,
all nodes are used, with activation multiplied by $1-p$.
\end{compactitem}

\section{Individual Regularization Behaviors}\label{sIndividual}

This section compares the behavior of each strategy.
We first conducted both experiments without regularization, achieving accuracies of
$54.02\pm 0.84\%$, $41.47\pm 2.85\%$, respectively.
Then we plot in Figure \ref{fIndividual} learning curves 
when each regularization strategy is applied individually.
We report training and validation accuracies through out this paper.
The main findings are as follows.

\begin{itemize}
\vspace{-.1cm}
\item Penalizing $\ell_2$-norm of weights helps generalization; the effect
depends largely on the size of training set. 
Experiment I contains 7,000 training samples and the improvement is 6.98\%; 
Experiment II contains more than 150k samples, and the improvement is only
2.07\%. Such results are consistent with other machine learning models.
\item  Penalizing $\ell_2$-norm of embeddings unexpectedly helps optimization (improves training accuracy). 
One plausible explanation is that since embeddings are trained on a large corpus by unsupervised methods, 
 they tend to settle down to large values and may not perfectly agree with the tasks of interest. 
$\ell_2$ penalty pushes the embeddings towards
small values and thus helps optimization. Regarding validation accuracy, Experiment I is improved by 6.89\%, whereas Experiment II has no significant difference.

\vspace{-.1cm}

\item  Re-embedding words does not improve
generalization. Particularly, in Experiment II, the ultimate accuracy is improved by 0.44, which is 
not large. Further, too much penalty hurts the models in both experiments.
In  the limit $\lambda_\text{reembed}$ to infinity, re-embedding words is mathematically 
equivalent to using embeddings as surface features, that is, freezing embeddings. Such strategy is sometimes applied in the literature like \newcite{cnn3}, but is not favorable
as suggested by the experiment.

\vspace{-.1cm}

\item Dropout helps generalization. 
Under the best settings, the eventual accuracy is improved by 3.12\% and 1.76\%, respectively.
In our experiments, dropout alone is not as useful as $\ell_2$ penalty. However, 
other studies report that dropout is very effective \cite{deepRNN}. Our results are
not consistent; different dimensionality may contribute to this disagreement, but more experiments are needed to confirm the hypothesis.
\end{itemize}

\begin{figure*}[!t]
\begin{center}
${\includegraphics[width=.43\textwidth]{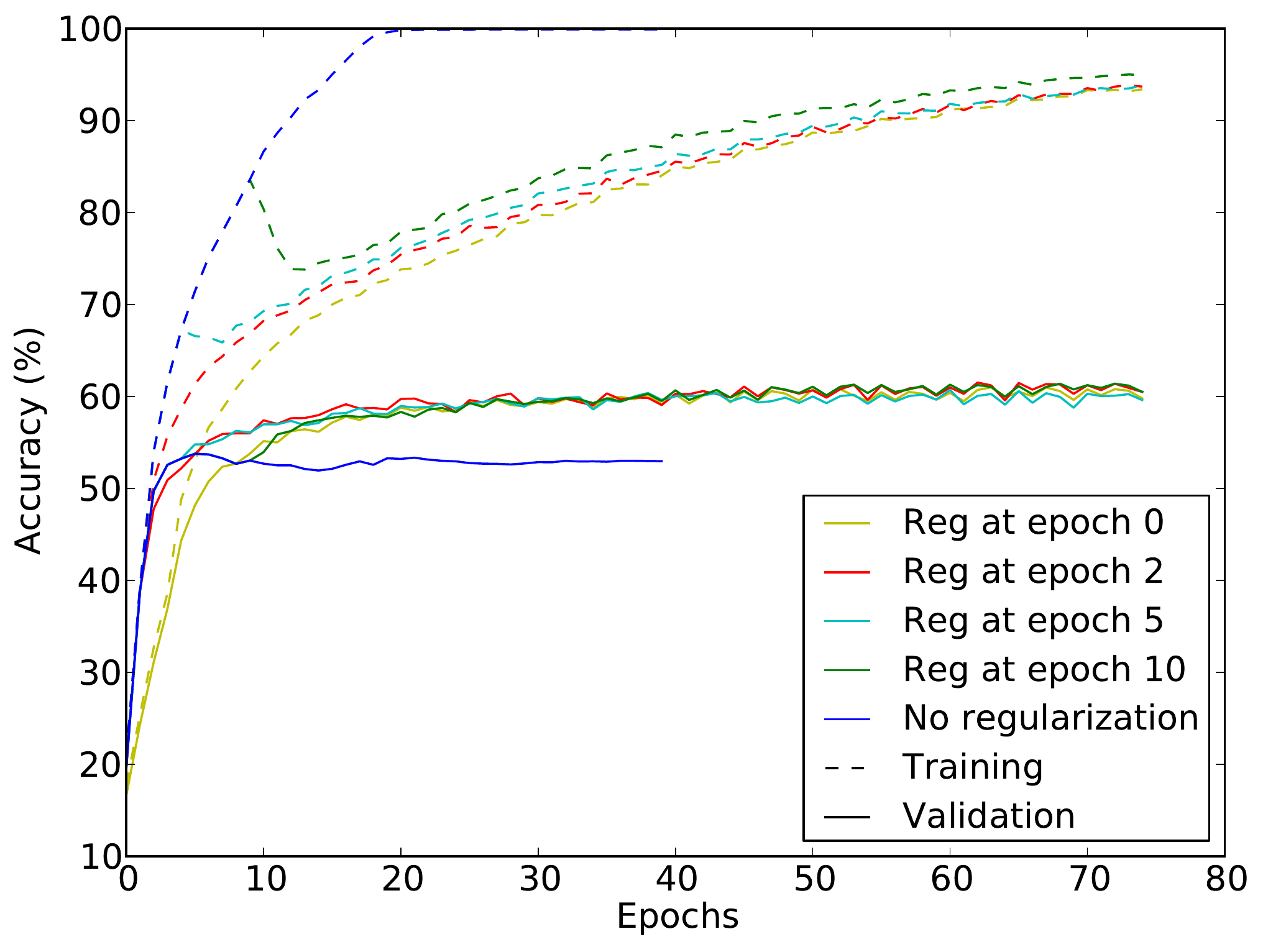}\atop \text{(a) Incrementally penalizing $\ell_2$-norm of weights.}}$\ \ \ \ \ \ 
${\includegraphics[width=.43\textwidth]{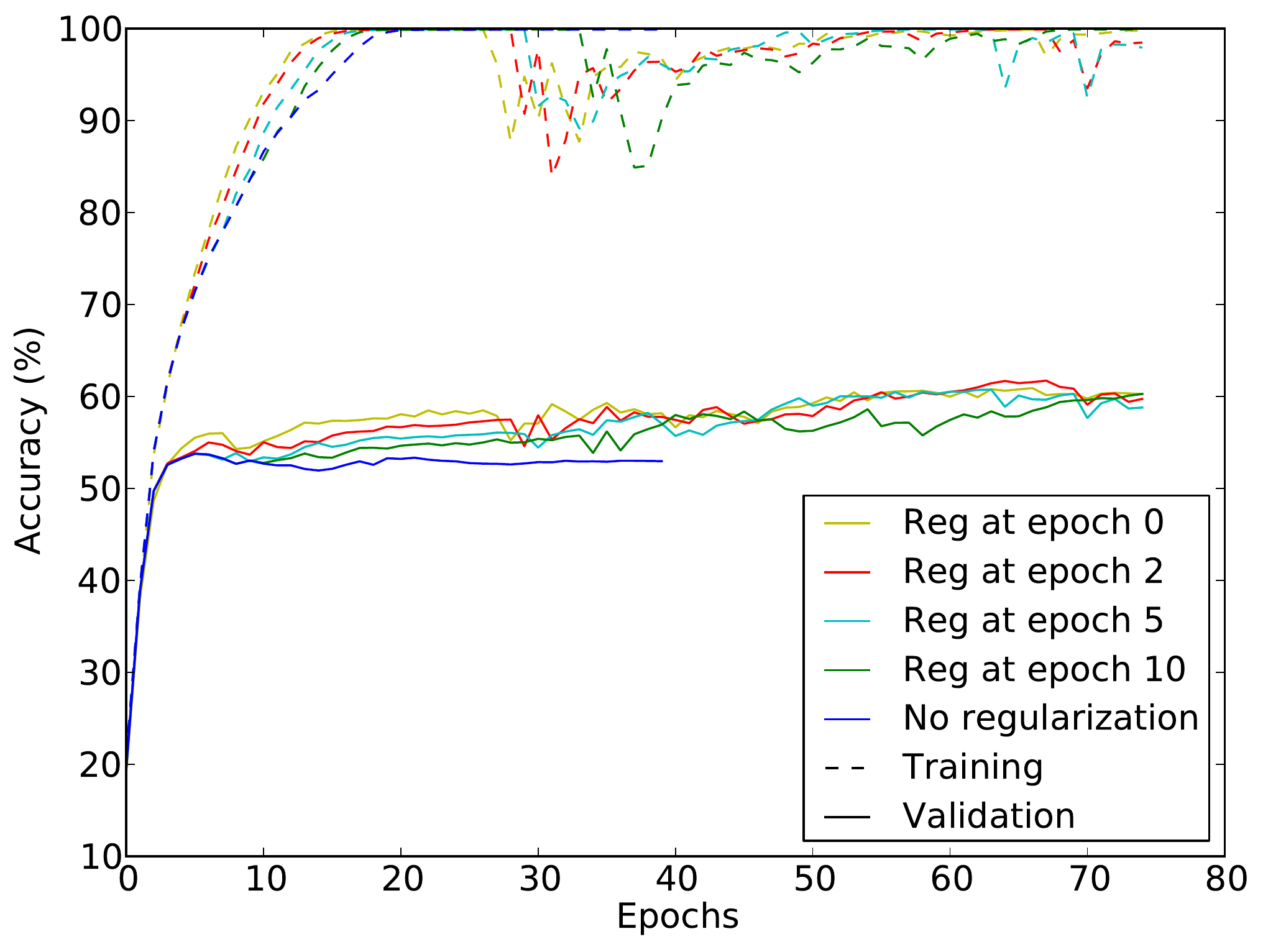}\atop \text{(b) Incrementally penalizing $\ell_2$-norm of biases.}}$\\
\vspace{-.2cm}
\caption{Tuning hyperparameters incrementally in Experiment I. Penalty is added at epochs 0, 2, 5, 10, respectively.
We chose the coefficients yielding the best performance in Figure \ref{fIndividual}. The controlled trial (no regularization) is early stopped
because the accuracy has already decreased.} 
\vspace{-.5cm}
\label{fIncremental}
\end{center}
\end{figure*}

\section{Incremental Hyperparameter Tuning}\label{sIncremental}
\vspace{-.2cm}
The above experiments show that regularization generally helps prevent overfitting.
To pursue the best performance, we need to
try out different hyperparameters through validation. 
Unfortunately, training deep neural networks is time-consuming, preventing
full grid search from being a practical technique. 
Things will get easier if we can incrementally tune hyperparameters, that is, to train the model without regularization first, and then add penalty.

In this section, we study whether $\ell_2$ penalty of weights and embeddings can be tuned incrementally. We exclude the dropout strategy because its does not make much sense
to incrementally drop out hidden units. 
Besides, from this section, we only focus on Experiment I 
due to time and space limit. 

Before continuing, we may envision several possibilities  
on how regularization works.
\begin{compactitem}
\item (On initial effects) As $\ell_2$-norm prevents parameters from growing large, adding it at early stages may cause parameters settling down to local optima. 
If this is the case, delayed penalty 
would help parameters get over local optima, leading to better performance.
\item (On eventual effects)
$\ell_2$ penalty lifts error surface of large weights.
Adding such penalty may cause parameters settling down
to (a) almost the same catchment basin, or (b) different basins.
In case (a), when the penalty is added does not matter much. In case (b), however,
it makes difference, 
because parameters would have already gravitated to catchment basins of 
larger values before regularization is added, which means incremental hyperparameter tuning
would be ineffective.

\end{compactitem}

To verify the above conjectures, we design four settings: adding penalty (1) at the beginning,
(2) before overfitting at epoch 2, (3) at peak performance (epoch 5), and (4) after overfitting (validation accuracy drops) at epoch 10.

Figure \ref{fIncremental} plots the learning curves regarding penalizing weights and embeddings, respectively; baseline (without regularization) is also included.

For both weights and embeddings, all settings yield similar ultimate validation accuracies. 
This shows $\ell_2$ regularization mainly serves as a ``local'' effect---it changes the error surface, but parameters tend to 
settle down to a same catchment basin.
We notice a recent report also shows local optima
may not play an important role in training neural networks, 
if the effect of parameter symmetry is ruled out \cite{localoptima}.

We also observe that 
regularization helps generalization as soon as it is added (Figure \ref{fIncremental}a), and that regularizing embeddings helps optimization also right after the penalty is applied
(Figure \ref{fIncremental}b).

\begin{table}[!t]
\centering
\vspace{-.2cm}
\resizebox{.4\textwidth}{!}{
\begin{tabular}{r||cccc}
\hline
     &  \multicolumn{4}{c}{$\lambda_\text{W}$}\\
  $\lambda_{\text{embed}}$ &      0		 &	$10^{-4}$	&	$3\!\cdot\!10^{-4}$	&	$10^{-3}$    \\
\hline
 $0\ \ \ \ \ $	       	&	54.02&	57.88&	59.96 &	61.00\\	
 $10^{-5}$	    &	54.94&	57.82&	60.68&	62.05\\
 $3\!\cdot\! 10^{-5}$	    & 	55.68&	61.02&	\textbf{64.00}&	\textbf{63.15}\\
 $10^{-4}$	    & 	60.91&	\textbf{64.00}&	\textbf{63.07}&	60.56\\
 $3\!\cdot\!10^{-4}$	    & 	58.92&	61.33&	59.85&	42.93\\
 $10^{-3}$	&	54.77&	56.43&	54.05&	16.50\\
\hline
\end{tabular}
}
\vspace{-.2cm}
\caption{Accuracy in percentage when we combine $\ell_2$-norm of weights and embeddings
(Experiment I).
Bold numbers are among highest accuracies (greater than peak performance minus 1.5 times standard deviation, i.e., 1.26 in percentage).}
\label{tCombineWE}

\centering
\resizebox{.42\textwidth}{!}{
\begin{tabular}{c|ccc|ccc}
\hline

                     & \multicolumn{3}{c|}{$\lambda_W$} & \multicolumn{3}{c}{$\lambda_\text{embed}$}\\
                     
$p$ 
        &\!\!$10^{\text{--}4}$  &\!\!\! $3\!\cdot\!10^{\text{--}4}$\!\!\!  & $10^{\text{--}3}$ &$10^{\text{--}5}$  &\!\!\! $3\!\cdot\!10^{\text{--}5}$\!\!\! &\!\!\! $10^{\text{--}4}$\!\!\!  \\
\hline  
\!\!0\!\!&\!\!\!57.88&\!\!\!\textbf{59.96} &\!\!\!\textbf{61.00}&\!\!54.94&\!\!\!55.68&\!\!\!\textbf{60.91}\!\!\!\\
\!\!1/6\!\!	&\!\!\! 58.36&\!\!\!59.36 &\!\!\!43.42&\!\!\!    58.49&\!\!\!59.59 &\!\!\!\textbf{60.00}\!\!\!\\
\!\!2/6\!\!	&\!\!\! 58.22&\!\!\! \textbf{60.00} &\!\!\!16.60	&\!\!\!  59.34&\!\!\!\textbf{60.08}&\!\!\! 59.61\!\!\!\\
\!\!3/6\!\!	&\!\!\! 58.63&\!\!\! 59.73 &\!\!\!16.60	&\!\!\!  59.59&\!\!\!\textbf{59.98}&\!\!\! 58.82\!\!\!\\
\!\!4/6\!\!	&\!\!\! 56.43&\!\!\! 54.63&\!\!\!16.60	&\!\!\!  56.76&\!\!\!59.19  &\!\!\! 56.64\!\!\!\\
\!\!5/6\!\!	&\!\!\! 38.07&\!\!\! 16.60&\!\!\!16.60	&\!\!\!  49.79&\!\!\!53.63 &\!\!\! 49.75\!\!\!\\
\hline
\end{tabular}}
\vspace{-.2cm}
\caption{Combining $\ell_2$ regularization and dropout.
Left: connectional weights. Right: embeddings.
($p$ refers to the dropout rate.)
}
\vspace{-.3cm}
\label{tCombineDropout}

\end{table}

\vspace{-.2cm}
\section{Combination of Regularizations}\label{sCombination}
\vspace{-.2cm}
We are further curious about the behaviors when different regularization methods are combined.

Table \ref{tCombineWE} shows that combining $\ell_2$-norm of weights and embeddings
results in a further accuracy improvement of 3--4 percents from applying either single one of them. In a certain range of coefficients, 
weights and embeddings are complementary: given one hyperparameter, we can tune the other to 
achieve a result among highest ones.

Such compensation is also observed in penalizing $\ell_2$-norm versus dropout (Table \ref{tCombineDropout})---although the peak performance is obtained by pure $\ell_2$ regularization,
applying dropout with small $\ell_2$ penalty also achieves a similar accuracy. 
The dropout rate is not very sensitive, provided it is small.

\vspace{-.2cm}
\section{Discussion}\label{sDiscussion}
\vspace{-.2cm}
In this paper, we systematically compared four regularization
strategies for embedding-based neural networks in NLP.
Based on the experimental results, we answer our research questions as follows.
(1) Regularization methods (except re-embedding words) basically help generalization.
Penalizing $\ell_2$-norm of embeddings unexpectedly helps optimization as well. 
Regularization performance depends largely on the dataset's size. (2) $\ell_2$ penalty mainly acts as a local effect; 
hyperparameters can be tuned incrementally. 
(3) Combining $\ell_2$-norm of weights
and biases (dropout and $\ell_2$ penalty) further improves  generalization; their coefficients are mostly complementary within a certain range.
These empirical results of regularization strategies shed some light
on tuning neural models for NLP.

\vspace{-.2cm}
\section*{Acknowledgments}\label{sAcknowledgment}
\vspace{-.2cm}
This research is supported by the National Basic Research Program of China (the 973 Program) under Grant No.\!\!\! 2015CB352201 and the National Natural Science Foundation of China under Grant No. 61232015. We would also like to thank Hao Jia and Ran Jia.


\newpage
\bibliographystyle{acl}
\bibliography{dl,dl2}

\end{document}